# A Unified Interpretable Semantic Framework for Multi-Modal Fiber Sensing Under Data Scarcity

Khouloud Abdelli
Advanced Photonics Department
Nokia Bell Labs
Stuttgart, Germany
khouloud.abdelli@nokia.com

*Abstract*—**We propose a unified interpretable semantic framework for multi-modal fiber sensing under data scarcity. The framework maps heterogeneous measurements from state-of-polarization (SOP) monitoring, distributed acoustic sensing (DAS), and optical time-domain reflectometry (OTDR) into a shared set of physics-grounded semantic descriptors capturing impulsiveness, persistence, structural characteristics, and energy concentration. Inference is performed through reasoning over semantic class profiles, optionally assisted by retrieval of semantically similar examples to resolve ambiguity and support adaptation without model retraining. Experimental validation on real SOP, DAS, and OTDR datasets demonstrates robust performance across sensing modalities, achieving accuracies of 95.0%, 91.1%, and 94.6%, respectively, while providing transparent and physically interpretable decisions. The proposed framework establishes a unified and data-efficient approach for optical fiber sensing in evolving and data-scarce deployment environments.**

## I. INTRODUCTION

Optical fiber sensing technologies are increasingly deployed in communication network monitoring, infrastructure surveillance, industrial automation, and environmental sensing applications [1, 2]. Diverse sensing modalities, including state-of-polarization (SOP) monitoring, distributed acoustic sensing (DAS), and optical time-domain reflectometry (OTDR), enable the detection and classification of a wide range of physical events and faults along optical fibers [3, 4]. Accurate event recognition is essential for fault diagnosis, network automation, condition monitoring, and autonomous system operation.

Despite significant progress in machine learning for optical sensing, reliable event classification remains challenging in practical deployments. Measurements are often affected by noise, environmental variability, and changing operating conditions, while labeled data are typically scarce, imbalanced, and expensive to acquire. Existing approaches predominantly rely on supervised models trained independently for each sensing modality. Although these methods often achieve excellent performance under controlled conditions, they generally require large labeled datasets and periodic retraining to maintain performance under distribution shifts. Furthermore, many models operate as black boxes, providing limited insight into the physical mechanisms underlying their decisions and reducing operational trust in critical monitoring applications.

A key observation motivates this work: although SOP, DAS, and OTDR measure fundamentally different physical quantities, many sensing events exhibit common behavioral characteristics, including impulsiveness, persistence, structural signatures, complexity, and energy concentration. These attributes are fundamentally physical rather than modality-specific, suggesting that heterogeneous sensing systems may share a common semantic representation. Domain experts routinely exploit such qualitative descriptions when interpreting sensing measurements; however, this knowledge is rarely formalized and systematically integrated into automated inference frameworks.

In this paper, we propose a unified interpretable semantic framework for multi-modal fiber sensing under data scarcity. The proposed approach maps heterogeneous sensing measurements into a shared set of physics-grounded semantic descriptors and performs inference through reasoning over semantic class profiles, optionally assisted by retrieval of semantically similar examples when ambiguity arises. The framework enables zero-shot and few-shot classification while providing physically interpretable decisions and supporting adaptation through retrieval without model retraining. Experimental validation is performed on real SOP, DAS, and OTDR datasets comprising mechanical disturbances, vibration events, and representative fiber faults. The proposed framework achieves accuracies of 95.0%, 91.1%, and 94.6% across the three sensing modalities, respectively, demonstrating robust performance and consistent cross-modal generalization under data-scarce conditions. The main contributions of this work are summarized as follows: (i) a unified semantic representation that maps heterogeneous optical sensing measurements into a shared set of physics-grounded descriptors; (ii) an interpretable semantic inference framework enabling zero-shot and few-shot event classification without model retraining; (iii) a retrieval-assisted adaptation mechanism for resolving ambiguous or overlapping event signatures; and (iv) comprehensive experimental validation across SOP, DAS, and OTDR datasets acquired under realistic deployment conditions.

## II. UNIFIED SEMANTIC INFERENCE FRAMEWORK

Fig. 1 illustrates the proposed framework, which enables interpretable and data-efficient event classification by decoupling modality-specific signal processing from modality-independent inference through a shared semantic representation of signal behavior. The framework is organized into four stages: (1) modality-specific feature extraction, (2) feature-to-semantic abstraction, (3) semantic inference, and (4) retrieval-assisted refinement and adaptation. While feature extraction and semantic calibration depend on the sensing modality, all subsequent operations are performed within a unified semantic representation, enabling consistent reasoning across heterogeneous sensing systems.

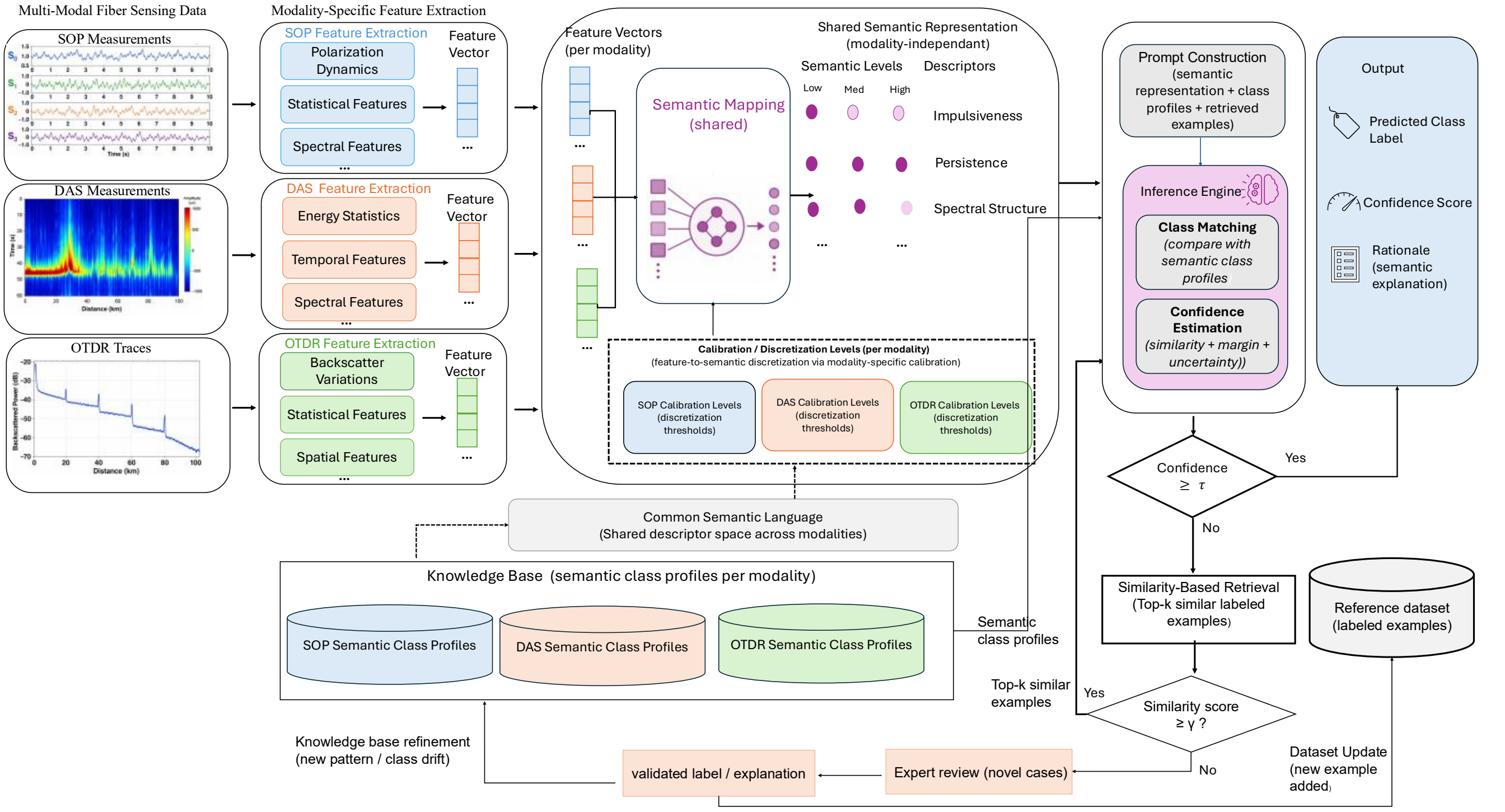

*Fig. 1. Unified semantic framework for multi-modal fiber sensing. Modality-specific measurements are transformed into a shared semantic representation through feature extraction and semantic mapping, followed by retrieval-assisted semantic reasoning and knowledge-base adaptation.*

In the first stage, raw measurements from SOP, DAS, and OTDR systems are processed through modality-specific pipelines to extract numerical descriptors that capture the underlying sensing physics. For SOP measurements, the extracted descriptors characterize polarization dynamics, temporal evolution, and spectral fluctuations. DAS features describe vibration energy distribution, temporal behavior, and frequency-domain characteristics, while OTDR descriptors quantify reflection peaks, attenuation profiles, and spatial discontinuities associated with fiber faults. Although these descriptors differ substantially across sensing modalities, they provide complementary descriptions of signal behavior that are subsequently mapped into a common semantic language composed of physics-grounded attributes, including impulsiveness, persistence, spectral or structural profile, complexity, and energy concentration. This transformation enables heterogeneous sensing measurements to be represented within a modality-independent semantic space suitable for cross-modal reasoning. Table I summarizes the semantic descriptor mappings adopted for each sensing modality.

To support robust inference and improve resilience to measurement variability and noise, the continuous-valued semantic descriptors are discretized into qualitative levels (Low, Medium, and High) using modality-specific calibration thresholds derived from feature distributions and domain knowledge. The resulting semantic representation is combined with semantic class profiles stored in a knowledge base, where each profile encodes the expected pattern of semantic descriptors associated with a particular event class. These class profiles provide structured prior knowledge that guides the inference process while remaining independent of the underlying sensing modality.

TABLE I: Unified Semantic Descriptor Mappings Across SOP, DAS, and OTDR Modalities

| **Semantic Descriptor** | **Physical Meaning** | **SOP** | **DAS** | **OTDR** |
|---|---|---|---|---|
| **Impulsiveness** | Suddenness and sharpness of localized signal variations | Polarization-derivative kurtosis, SOP velocity peaks | Peak amplitude, signal kurtosis | Normalized gradient, peak-to-mean ratio |
| **Persistence** | Duration or spatial extent of event signatures | SOP motion duration, sustained SOP activity | Temporal spread, event duration | Fault spatial extent, event width |
| **Spectral / Structural Profile** | Frequency composition or attenuation/reflection structure | Spectral spread, dominant frequency | Band-energy ratios, spectral score | Reflectance, attenuation slope |
| **Complexity** | Irregularity and structural variability | Angular-speed variance, trajectory entropy | Scalogram entropy | Trace entropy |
| **Energy Concentration** | Degree to which energy is localized or distributed | SOP excursion amplitude | Energy concentration ratio | Dynamic range, localized reflection energy |

Inference is subsequently performed entirely within the semantic domain. The semantic representation of an input sample is combined with the corresponding semantic class profiles to construct the inference context. A semantic reasoning engine evaluates the consistency between the observed descriptor pattern and candidate class profiles to generate a predicted class label, confidence score, and interpretable rationale linking the decision to physically meaningful signal characteristics.

When the confidence score falls below a predefined threshold, a retrieval mechanism is activated to identify a small number of semantically similar labeled examples from a reference database. These examples provide instance-level evidence that complements the semantic class profiles and improves discrimination between overlapping classes. If uncertainty persists or previously unseen patterns are encountered, expert review may be invoked to update both the semantic knowledge base and the retrieval database. Consequently, the same reasoning framework operates unchanged across SOP, DAS, and OTDR despite their fundamentally different sensing physics, enabling unified, interpretable, and adaptive cross-modal fiber sensing under data-scarce deployment conditions.

## III. DATASETS AND SEMANTIC CONFIGURATION

### *A. SOP Disturbance Dataset*

SOP validation is performed on a 2600-km ROADM-based WDM transmission system operating under fully loaded C-band conditions [5]. SOP measurements are acquired from a 62-Gbaud DP-QPSK channel under test using millisecond-resolution Stokes monitoring. Mechanical perturbations are generated using an Arduino-controlled actuation platform and include fiber bending, shaking, impact, and up-down motion, together with a normal operating condition, resulting in a five-class classification problem. To improve robustness and reduce polarization-state bias, a polarization scrambler randomizes the initial SOP before each acquisition. For semantic inference, SOP evolution is represented using quaternion-based polarization trajectories. Numerical descriptors including polarization-derivative kurtosis, SOP angular-speed statistics, spectral spread, dominant frequency, and excursion amplitude are mapped into the semantic dimensions of impulsiveness, persistence, spectral profile, complexity, and energy concentration.

### *B. DAS Field-Trial Dataset*

DAS validation uses field measurements acquired on a deployed 57-km SSMF link instrumented using a MIMO distributed fiber sensing system with a 4.1-m gauge length and a 2.6-ms probing interval [6]. Three classes are considered: background activity, jackhammer operation, and excavator digging. Continuous wavelet transform analysis is applied to distributed phase measurements, and each disturbance is represented as a 96×96 scalogram extracted around the dominant energy region. Semantic descriptors are derived from the time-frequency representation using peak amplitude, band-energy ratios, entropy measures, temporal spread, and concentration metrics, yielding semantic dimensions describing impulsiveness, persistence, spectral structure, complexity, and energy concentration.

### *C. OTDR Fault Dataset*

OTDR validation employs an experimental fault dataset containing representative access-network anomalies including fiber tapping, bad splice, dirty connector, bending events, reflector events, and fiber cuts. OTDR traces are acquired at 1650 nm using 10-ns pulses and represented as normalized 30-point power profiles together with auxiliary metadata including reflectance, insertion loss, and signal-to-noise ratio [3]. Semantic descriptors characterize abrupt power variations, spatial persistence, structural attenuation behavior, complexity, energy localization, and signal quality using features including reflectance, attenuation slope, gradient statistics, entropy measures, and peak-to-mean ratios.

### *D. Semantic Class Profiles*

Semantic class profiles define the expected descriptor patterns associated with each event type and provide structured prior knowledge for the reasoning engine. Rather than imposing rigid decision boundaries, the profiles encode dominant behavioral tendencies while allowing overlap between related classes to accommodate measurement variability and uncertainty. The semantic profiles used in this work are shown in Fig. 2.

## IV. EXPERIMENTAL RESULTS

The proposed framework was evaluated on three fundamentally different optical sensing modalities: (i) SOP disturbance classification using measurements acquired on a 2600-km ROADM-based WDM transmission system, (ii) field-trial DAS vibration monitoring on a deployed 57-km fiber link, and (iii) OTDR fault diagnosis across eight events. Across all experiments, modality-specific measurements were mapped into the unified semantic descriptor space defined in Table I. Semantic inference was performed through reasoning over semantic class profiles using an LLM-based reasoning engine implemented with GPT-4.1-mini, while retrieval-assisted refinement was activated whenever the confidence score fell below a predefined threshold.

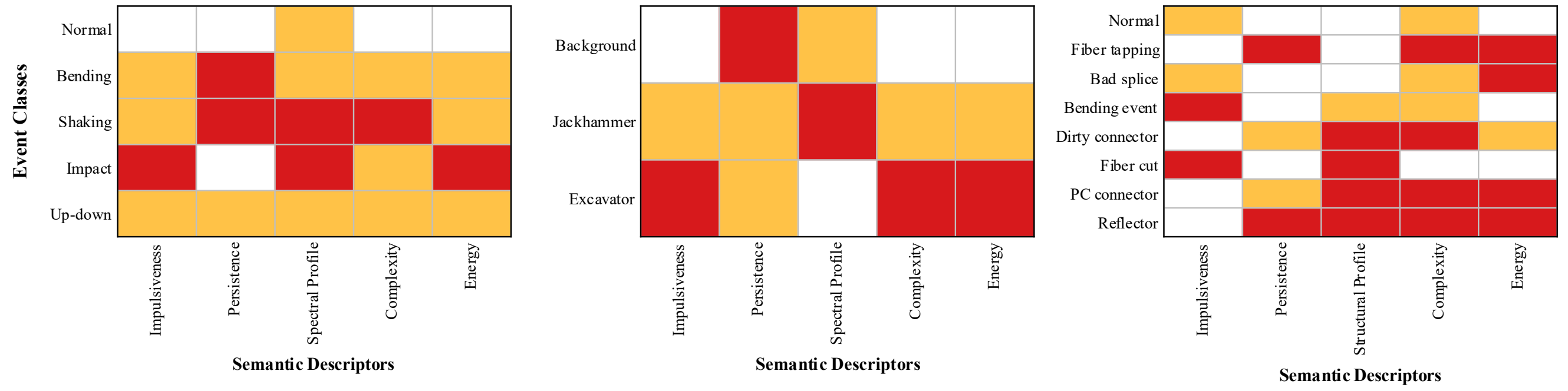

Fig. 2. Semantic class profiles across SOP, DAS, and OTDR sensing modalities represented using the proposed shared semantic descriptors.

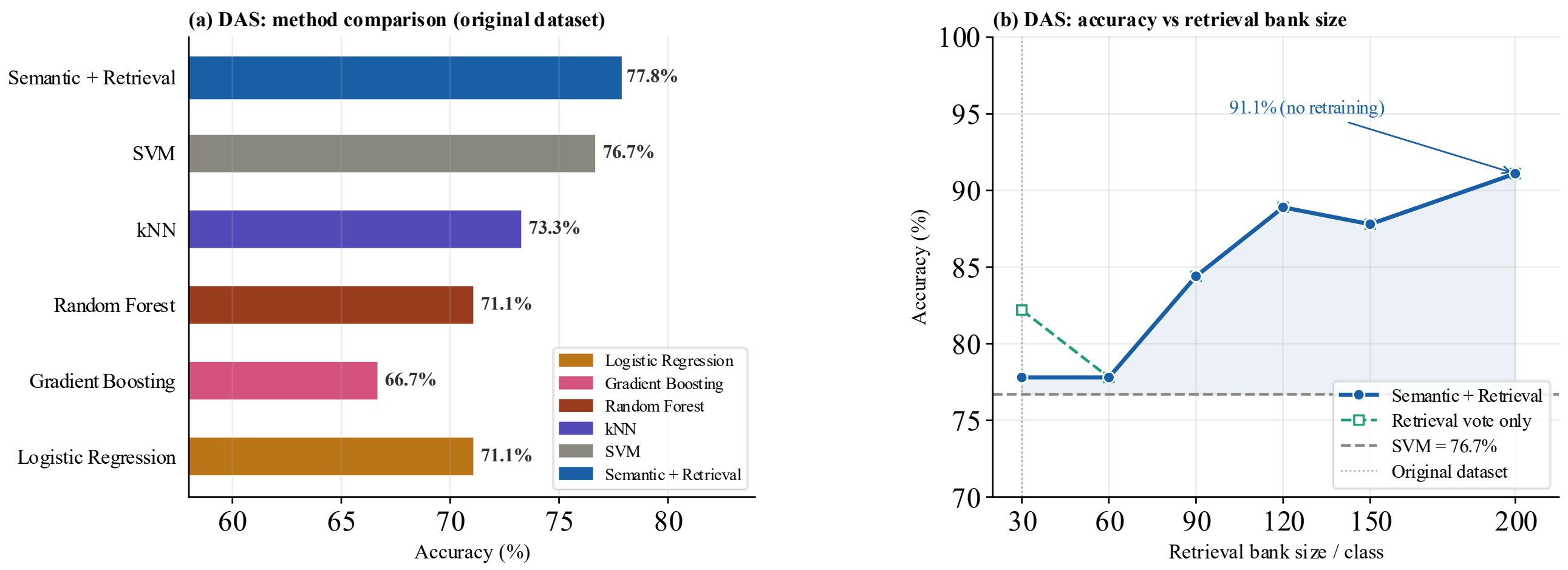

Fig. 4.(a) DAS classification performance under low-data conditions (~30 samples per class). (b) Effect of retrieval bank size on DAS classification accuracy without retraining.

## A. *SOP Disturbance Classification*

Figure 3 summarizes the obtained performance for SOP disturbance classification on the five-class dataset acquired using the 2600-km ROADM-based WDM transmission system. The proposed retrieval-enhanced semantic framework achieves an average classification accuracy of 95.0 ± 0.01% across five independent runs, outperforming both a Support Vector Machine (SVM) classifier (90.0%) and a Multilayer Perceptron (MLP) neural network (91.0%) trained using approximately 9,000 labeled samples under five-fold cross-validation.

The semantic-only configuration achieves 90.0% accuracy, demonstrating that the proposed semantic descriptors and class profiles alone capture the dominant characteristics of polarization disturbances despite the absence of supervised training or model optimization. Incorporating retrieval-assisted refinement provides an additional 5 percentage-point improvement, increasing performance to 95.0%.

The largest performance gains are observed for the shaking and impact disturbances, which exhibit partially overlapping temporal and spectral characteristics but differ in impulsiveness and energy concentration. In these cases, retrieved examples provide complementary instance-level evidence that refines the semantic reasoning process and improves discrimination between semantically similar classes.

A rule-based baseline relying exclusively on manually selected thresholds achieves only 76.0% accuracy, highlighting the limitations of fixed decision boundaries for complex polarization dynamics and motivating the use of reasoning over multiple interacting semantic descriptors rather than isolated features.

These results demonstrate that semantic inference can effectively characterize polarization disturbances while requiring neither large labeled datasets nor model retraining.

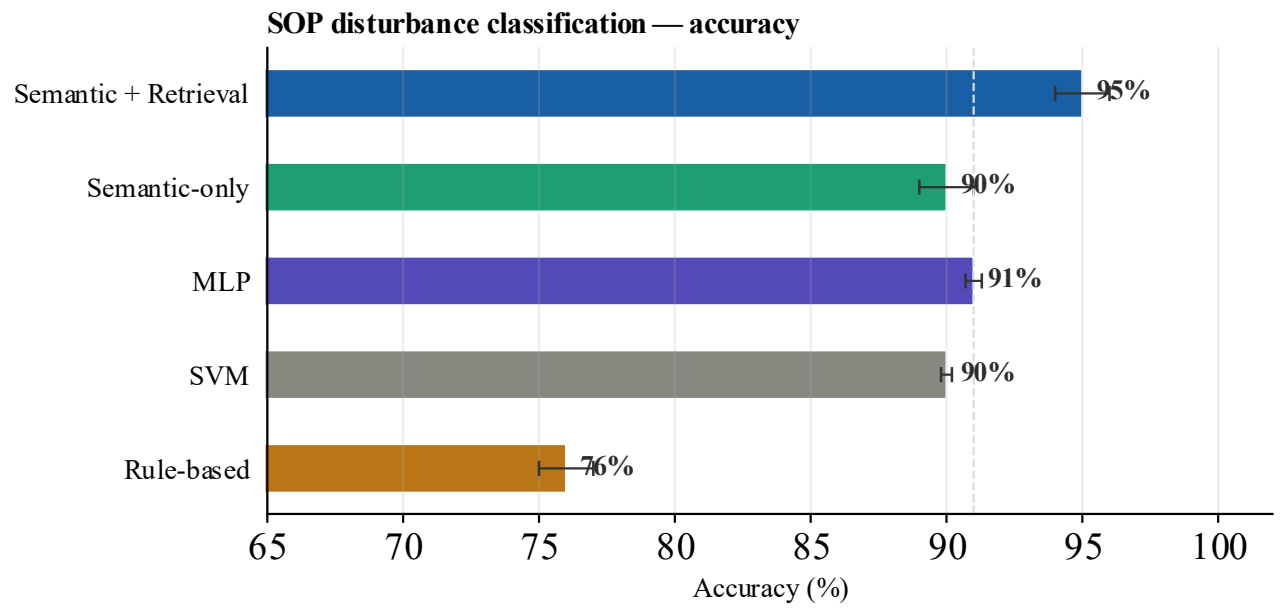

Fig. 3. Performance comparison for SOP disturbance classification. Retrieval-assisted semantic inference achieves the highest accuracy (95%) while requiring no retraining.

## B. *DAS Field-Trial Classification*

The proposed framework was evaluated on the field-trial DAS dataset containing background activity, jackhammer operation, and excavator digging events acquired on a deployed 57-km fiber link . Figure 4(a) compares the proposed approach against several conventional machine-learning baselines under the original low-data setting containing approximately 30 examples per class.

Under these strict data-scarce conditions, the retrieval-enhanced semantic framework achieves 77.8% classification accuracy without supervised training or parameter optimization, slightly outperforming a Support Vector Machine (SVM) classifier (76.7%) trained using sixteen handcrafted features and five-fold cross-validation. The proposed approach also exceeds k-nearest neighbors (73.3%), Random Forest (71.1%), Gradient Boosting (66.7%), and Logistic Regression (71.1%) baselines despite operating exclusively through semantic reasoning and retrieval.

The primary source of classification errors arises from ambiguity between background activity and jackhammer events, which exhibit partially overlapping temporal persistence and energy characteristics under realistic field conditions. In contrast, excavator activity exhibits stronger low-frequency energy concentration and higher impulsiveness, resulting in more distinctive semantic signatures.

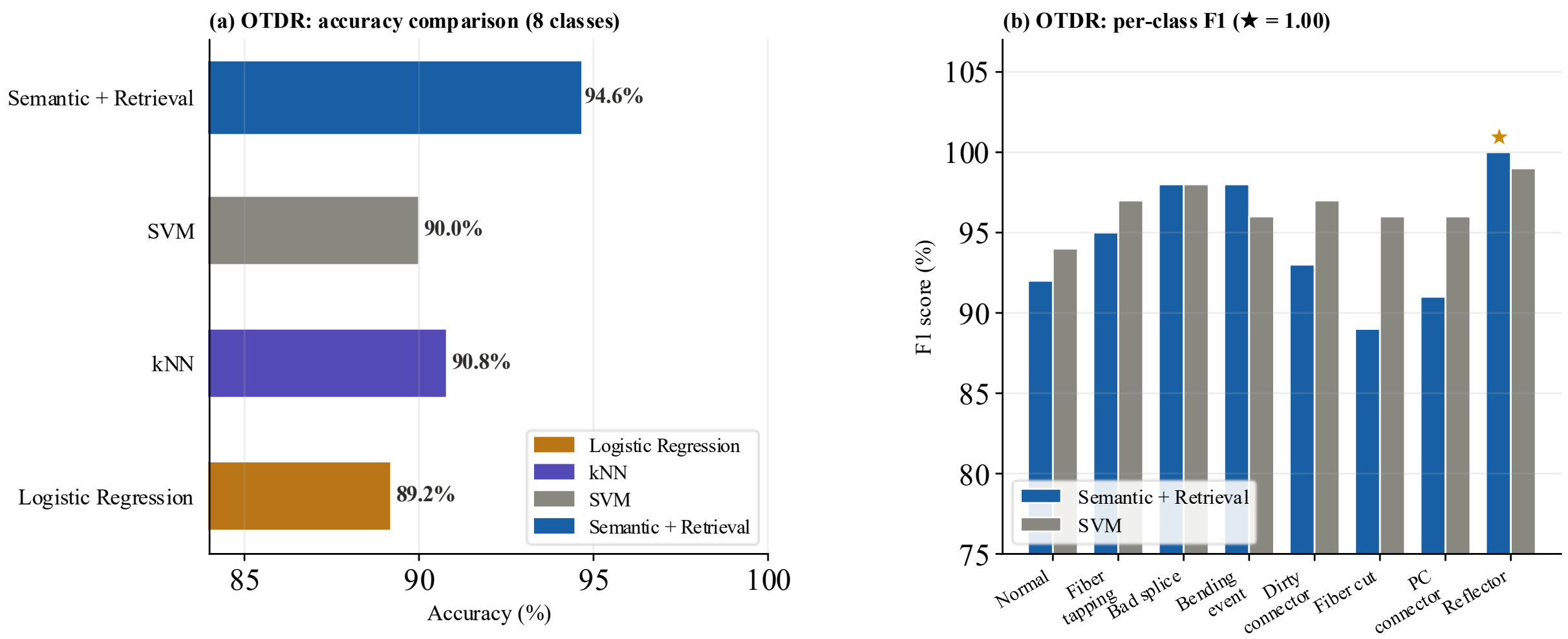

Fig. 5. OTDR fault classification results: (a) overall accuracy comparison and (b) per-class F1 scores for the proposed semantic framework and the SVM baseline.

A key advantage of the proposed framework is its ability to improve through knowledge accumulation rather than model retraining. As illustrated in Fig. 4(b), progressively expanding the retrieval database using additional labeled examples increases classification accuracy monotonically from 77.8% to 91.1% without modifying model parameters or updating the reasoning engine.

This behavior demonstrates the practical value of retrieval-assisted adaptation in operational sensing systems, where new examples become available gradually during deployment. Rather than retraining a supervised model, the proposed framework continuously improves by enriching the reference database used during semantic reasoning. Critically, this adaptation requires no ML engineering effort — an operator adding labeled examples during deployment automatically improves system performance without any model update cycle, making the framework operationally self-improving under real deployment conditions.

## C. *OTDR Fault Classification*

The proposed framework was further evaluated on the OTDR dataset containing eight operational states, including normal operation, fiber tapping, bad splice, bending events, dirty connectors, fiber cuts, PC connector faults, and reflector events. Figure 5(a) summarizes the obtained classification performance.

Using only semantic descriptors and retrieval-assisted reasoning, the proposed framework achieves an overall accuracy of 94.6% without supervised training, outperforming both k-nearest neighbors (kNN) (90.8%) and Logistic Regression (89.2%) baselines and exceeding the supervised Support Vector Machine (SVM) classifier (90.0%) trained on the same feature set.

Figure 5(b) further illustrates the per-class F1 scores obtained by the semantic framework and the SVM baseline. Several fault classes exhibit highly distinctive semantic signatures. In particular, Reflector events achieve perfect classification (F1 = 1.00) owing to their unique combination of strong structural response and highly concentrated energy localization. Similarly, Bad Splice and Bending Event classes achieve F1 scores of approximately 98%, reflecting their characteristic attenuation and structural patterns.

These results demonstrate that semantic reasoning can effectively capture the dominant physical mechanisms underlying OTDR faults while maintaining transparency and avoiding model retraining.

## D. *Interpretable Semantic Inference*

Beyond predictive performance, the proposed framework provides transparent outputs including a predicted class, confidence score, and semantic rationale supporting the decision. Figure 6 illustrates a representative OTDR example in which an input trace is transformed into semantic descriptors and subsequently classified as a Fiber Cut event.

The reasoning process identifies the event as exhibiting high impulsiveness, high structural profile, and high energy concentration, producing a semantic signature consistent with abrupt signal interruption and localized energy loss. Alternative explanations, such as dirty connectors or bad splices, are rejected owing to their weaker structural response and lower energy concentration.

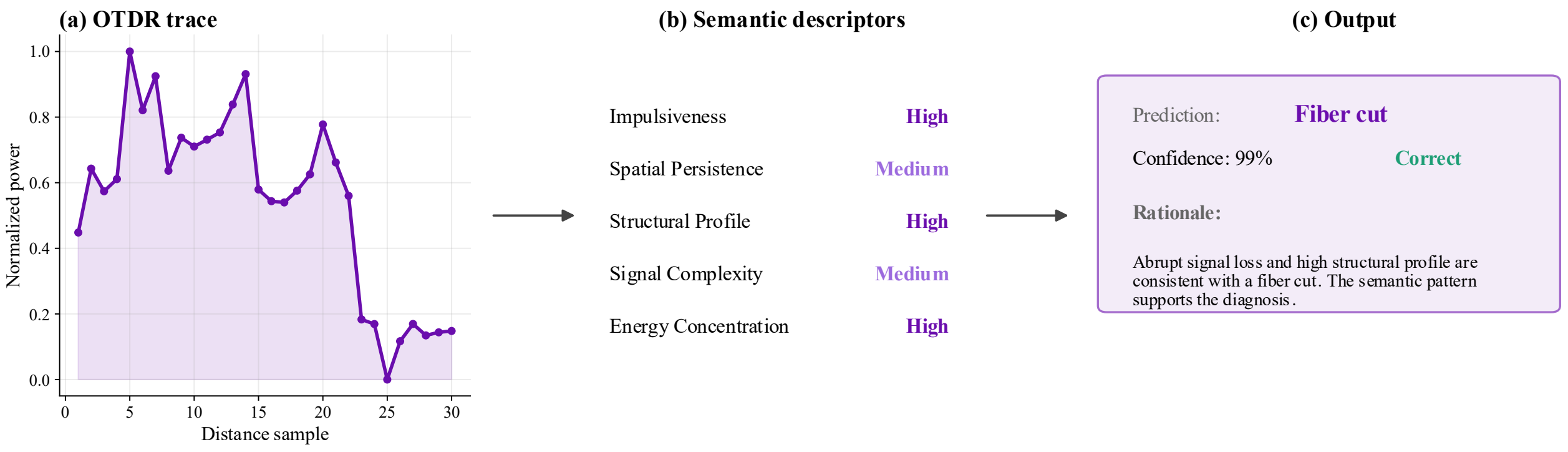

Fig. 6. Example of interpretable semantic inference for OTDR fault diagnosis. The measured trace is transformed into semantic descriptors and classified through semantic reasoning, producing a predicted class, confidence score, and human-readable rationale.

Unlike conventional black-box classifiers, the proposed framework enables operators to inspect both the prediction and the underlying semantic evidence, facilitating validation by domain experts and improving operational trust.

## V. DISCUSSION AND CONCLUSION

This work investigated semantic reasoning as a unified inference paradigm for heterogeneous optical fiber sensing modalities. Despite operating on fundamentally different measurements—including polarization trajectories, time-frequency vibration signatures, and reflectometry traces—the proposed framework employs a common semantic representation and a single reasoning engine across SOP, DAS, and OTDR sensing systems.

Experimental validation demonstrated competitive performance across all investigated modalities, achieving accuracies of 95.0%, 91.1%, and 94.6% for SOP, DAS, and OTDR, respectively, while providing transparent decisions through semantic explanations and confidence estimates.

The results suggest that many sensing events can be characterized using a small set of modality-independent behavioral descriptors related to impulsiveness, persistence, structural organization, complexity, and energy localization. This observation supports the central hypothesis of this work: heterogeneous optical sensing modalities share an underlying semantic structure that can serve as the basis for unified reasoning.

Furthermore, the retrieval-assisted adaptation mechanism enables continuous improvement through accumulation of reference examples rather than parameter optimization or model retraining. This property is particularly attractive for practical deployments where labeled data are scarce and sensing conditions evolve over time.

To the best of our knowledge, this work represents the first demonstration of a unified zero-shot semantic inference framework operating across SOP, DAS, and OTDR sensing modalities simultaneously, establishing the feasibility of interpretable, data-efficient, and adaptive cross-modal inference without supervised training or modality-specific model retraining.